\documentclass[10pt,twocolumn,letterpaper]{article}

\usepackage{cvpr}              

\usepackage{graphicx}
\usepackage{amsmath}
\usepackage{amssymb}
\usepackage{booktabs}

\usepackage{multirow}
\usepackage{bbm}
\usepackage[linesnumbered, ruled]{algorithm2e}
\usepackage{algpseudocode}

\usepackage[pagebackref,breaklinks,colorlinks]{hyperref}

\usepackage[capitalize]{cleveref}

\def\cvprPaperID{5805} 
\def\confName{CVPR}
\def\confYear{2023}

\def\ie{\emph{i.e.,}\xspace}

\def\market{{Market-1501}\xspace}
\def\msmt{{MSMT17}\xspace}
\def\cgc{{confidence-guided centroids}\xspace}
\def\cgl{{confidence-guided pseudo labels}\xspace}
\newcommand{\para}[1]{{\setlength{\parskip}{0.3em} \noindent \textbf {#1}}}

\usepackage{enumitem}
\newenvironment{tight_itemize}{
\begin{itemize}[leftmargin=20pt]
\setlength{\topsep}{0pt}
\setlength{\itemsep}{0pt}
\setlength{\parskip}{0pt}
\setlength{\parsep}{0pt}
}{\end{itemize}}

\begin{document}

\title{Confidence-guided Centroids for Unsupervised Person Re-Identification}

\author{Yunqi Miao\\
\small{University of Warwick}\\
\and
Jiankang Deng\\
Huawei\\
\and
Guiguang Ding\\
Tsinghua University\\
\and
Jungong Han\\
Aberystwyth University\\
}
\maketitle

\begin{abstract}
Unsupervised person re-identification (ReID) aims to train a feature extractor for identity retrieval without exploiting identity labels.
Due to the blind trust in imperfect clustering results, the learning is inevitably misled by unreliable pseudo labels.
Albeit the pseudo label refinement has been investigated by previous works, they generally leverage auxiliary information such as camera IDs and body part predictions. 
This work explores the internal characteristics of clusters to refine pseudo labels.
To this end, Confidence-Guided Centroids (CGC) are proposed to provide reliable cluster-wise prototypes for feature learning. 
Since samples with high confidence are exclusively involved in the formation of centroids, the identity information of low-confidence samples, \ie boundary samples, are NOT likely to contribute to the corresponding centroid.
Given the new centroids, current learning scheme, where samples are enforced to learn from their assigned centroids solely, is unwise.
To remedy the situation, we propose to use Confidence-Guided pseudo Label (CGL), which enables samples to approach not only the originally assigned centroid but other centroids that are potentially embedded with their identity information.
Empowered by confidence-guided centroids and labels, our method yields comparable performance with, or even outperforms, state-of-the-art pseudo label refinement works that largely leverage auxiliary information.
\end{abstract}

\section{Introduction}
\begin{figure}
\centering
\includegraphics[width=.98\linewidth]{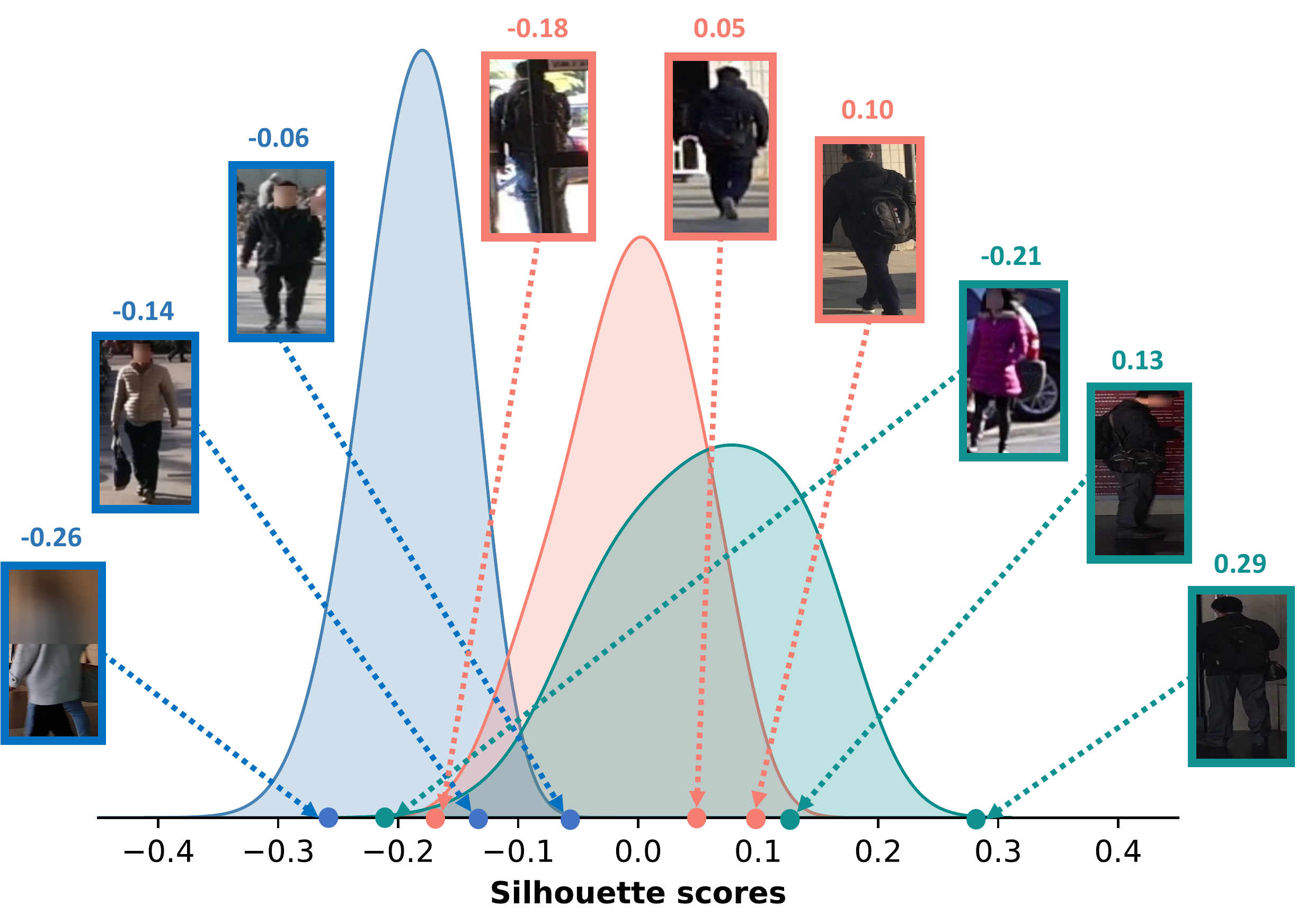}
\caption{Training samples (cluster ID = 1) and their silhouette scores at epoch 0 (blue), epoch 25 (orange), and epoch 50 (green) on \msmt~\cite{wei2018person}. Higher silhouette scores denote samples are clustered at higher confidence. \textbf{Best viewed in color}.}
\label{fig:sspic}
\vspace{-0.53cm}
\end{figure}
Person re-identification (ReID) aims to retrieve a person of interest across multiple cameras~\cite{ye2021deep,zheng2017wild,li2018hacnn}. Due to the label-free training manner, unsupervised person ReID methods have attracted increasing attention. Unsupervised ReID methods can be broadly categorized into two types: unsupervised domain adaptation (UDA) methods~\cite{zhong2019invariance,zhang2019self,ge2020self,dai2021idm,wu2022multi,he2022secret} and purely unsupervised learning (USL) methods~\cite{dai2021cluster,chen2021ice,cho2022part,wang2021camera,zhang2022implicit}. The former pre-trains a model on person-related datasets, \ie source domain, and fine-tunes it on ReID-related datasets, \ie target domain. Apart from requiring additional annotated labels, UDA methods are vulnerable to the large gap between the source domain and the target domain. In contrast, USL methods do not require any labeled data for training, which are more challenging but well fit real-world scenarios. In the paper, we focus on USL methods.

Existing USL methods generally follow a two-stage training scheme: 1) clustering, \ie obtaining the pseudo labels via a clustering algorithm such as DBSCAN~\cite{ester1996density}, and 2) network training, \ie optimizing the network in a ``supervised'' manner with assigned cluster IDs. Contrastive loss such as InfoNCE~\cite{ge2020self} or ClusterNCE~\cite{dai2021cluster} usually serves as training objectives. 
Due to the blind trust in imperfect clustering results, the learning is inevitably misled by unreliable pseudo labels, where multiple identities are merged into one cluster or samples of one person are assigned to multiple clusters.
Despite that some pseudo label refinement~\cite{wang2021camera,chen2021ice,zhang2021refining,zhang2022implicit,cho2022part} have been proposed, they generally leverage auxiliary information, such as camera IDs~\cite{wang2021camera,chen2021ice}, body part predictions~\cite{cho2022part}, and generated samples~\cite{zhang2022implicit}. 
Given the fact that such auxiliary information is not free in reality, refining pseudo labels by merely exploiting internal characteristics within samples, \ie the sample-wise clustering confidence, appears to be more valuable. 

To measure the sample-wise clustering confidence, \ie how well a sample fits its cluster, we employ a metric: silhouette score~\cite{rousseeuw1987silhouettes}. 
The score presents the ratio between intra-cluster distance and inter-cluster distance, which ranges from -1 to +1 (\textit{higher is better}).
To demonstrate the relationship between the clustering confidence and the silhouette score, we visualize silhouette scores of training samples of MSMT17~\cite{wei2018person} in Fig.~\ref{fig:sspic}. Samples are from the same cluster (cluster ID=1) but at different training epochs, \ie 0, 25, and 50, respectively. 
As training goes on, the clustering is gradually enhanced by involving more effective features and the more discriminative network.
At first images are grouped by coarse visual features, yet by identity-related information in the end.
Meanwhile, sample-wise silhouette scores continuously shift towards higher values during training.
Given this consistency, a conclusion can be drawn that, \textit{a higher silhouette score implies the sample better fits its cluster, i.e., being clustered at higher confidence.}
Previous learning schemes~\cite{dai2021cluster,zhang2022implicit} adopt all-sample based centroids, which are obtained by averaging features of all samples within the cluster, and enforce instances to approach such centroids.
However, our observation suggests that low-confidence samples either are poor in quality or belong to other identities. 
Features of such images will inevitably contaminate centroids regardless of the training stage. 
In light of this, we propose Confidence-Guided Centroids (CGC) to provide more reliable cluster-wise prototypes for feature learning.

Although the reliability of cluster centroids has been improved, the conventional one-hot labeling strategy aggravates a problem. 
Since high-confidence samples exclusively contribute to the formation of cluster centroids, the identity-related information of low-confidence samples can hardly be presented in the assigned centroid.
To illustrate the problem, an analysis is conducted on MSMT17~\cite{wei2018person}, where we intend to investigate how much identity information of low-confidence samples can be presented in their assigned centroids.
We found that, with the vanilla all-sample based cluster centroids, only 5.83$\%$ low-confidence samples have their identity information embedded in the assigned centroid at the beginning. Although the ratio gradually climbs to 17.19$\%$, a large proportion of low-confidence samples (over 80$\%$) still are pushed to ``wrong'' centroids.
Unfortunately, the ratio achieves 14.17$\%$ at most with \cgc.
Given the situation, the one-hot labeling strategy, which enforces samples to learn from the assigned centroid solely, is unwise. 
To address the problem, we propose to use \cgl (CGL), which encourages instances to approach not only the assigned confidence-guided centroid but also others where their identity information are potentially embedded.

In summary, our contributions are as follows:
\begin{tight_itemize}
    \item We propose Confidence-Guided Centroids (CGC) to provide cluster-wise prototypes for feature learning. The reliability of centroids is improved via filtering out low-confidence samples during formation.
    \item To overcome the problem that the identity information of low-confidence samples is rarely presented in their assigned centroids, we propose to use \cgl (CGL) during training. Apart from the originally assigned centroid, instances are also encouraged to approach other centroids where their identity information are potentially embedded.
    \item The proposed method only exploits internal characteristic for unsupervised person re-identification. Extensive experiments on benchmark datasets demonstrate that, our method yields better or comparable performances with state-of-the-art ones that largely leverage auxiliary information.
\end{tight_itemize}

\section{Related work}\label{sec:rw}
\para{Unsupervised Person ReID.}
The existing unsupervised person ReID methods are divided into two categories: a) Unsupervised Domain Adaptation (UDA) methods, which boost the performance by leveraging the knowledge transferred from the source domain~\cite{zhong2019invariance,zhang2019self,ge2020self,dai2021idm,wu2022multi,he2022secret}, and b) purely UnSupervised Learning (USL) methods, which do not require any identity labels during training~\cite{lin2020unsupervised,chen2021ice,zhang2021refining,dai2021cluster,zhang2022implicit,cho2022part}. 
Since UDA methods are highly prone to be affected by the large gap between the source domain and the target domain, they are hardly applicable to real-world scenarios~\cite{ye2021deep,lin2021unsupervised}. In the paper, we focus on USL methods. 

Generally, USL methods exploit pseudo labels, instead of actual identity labels, as the guidance during training. Pseudo labels can be generated either by the image similarity~\cite{lin2020unsupervised,wang2020unsupervised} or clustering algorithms~\cite{lin2019bottom,zeng2020hierarchical,dai2021cluster,zhang2022implicit}. 
Specifically, SSL~\cite{lin2020unsupervised} and MMCT~\cite{wang2020unsupervised} formulate unsupervised person ReID as a classification task and predict pseudo labels based on the image similarity. In terms of clustering-based methods, BUC~\cite{lin2019bottom} and HCT~\cite{zeng2020hierarchical} employ the bottom-up clustering scheme to gradually merge similar individual samples into clusters. Recently, Cluster-Contrast~\cite{dai2021cluster} adopts a contrastive learning scheme, which initializes, updates, and performs contrastive loss computation at the cluster level. 
However, clustering-based methods are generally sensitive to the pseudo label noise brought by imperfect clustering results. 

\para{Noise Reduction of Pseudo Label.}
Recently, how to handle noise pseudo labels in clustering-based methods has become a research hotspot.
Specifically, SpCL~\cite{ge2020self} employs a self-paced learning scheme to gradually obtain more reliable clusters for the pseudo label refinement. 
CAP~\cite{wang2021camera} splits each cluster into multiple proxies according to camera IDs. Such camera-aware proxies eliminate the pseudo label noise brought by varying viewing points.
ICE~\cite{chen2021ice} alleviates the label noise by enhancing the consistency between augmented and original instances.
RLCC~\cite{zhang2021refining} refines pseudo labels with the clustering consensus, which encourages the consistency between cluster results of two consecutive iterations.
PPLR~\cite{cho2022part} employs the complementary relationship between reliable features of human global and body parts for the pseudo label refinement.
ISE~\cite{zhang2022implicit} generates boundary samples from actual samples and their neighboring clusters. The discriminability of the network is improved by enforcing generated samples to be correctly classified.

Unlike the above methods, this work explores whether internal characteristics can facilitate the pseudo label refinement. In the paper, we investigate the sample-wise clustering confidence, which describes how well a sample fits its cluster. With such criteria, we propose to employ better cluster centroids and pseudo labels for feature learning. 

\begin{figure*}
\centering
\includegraphics[width=.96\linewidth]{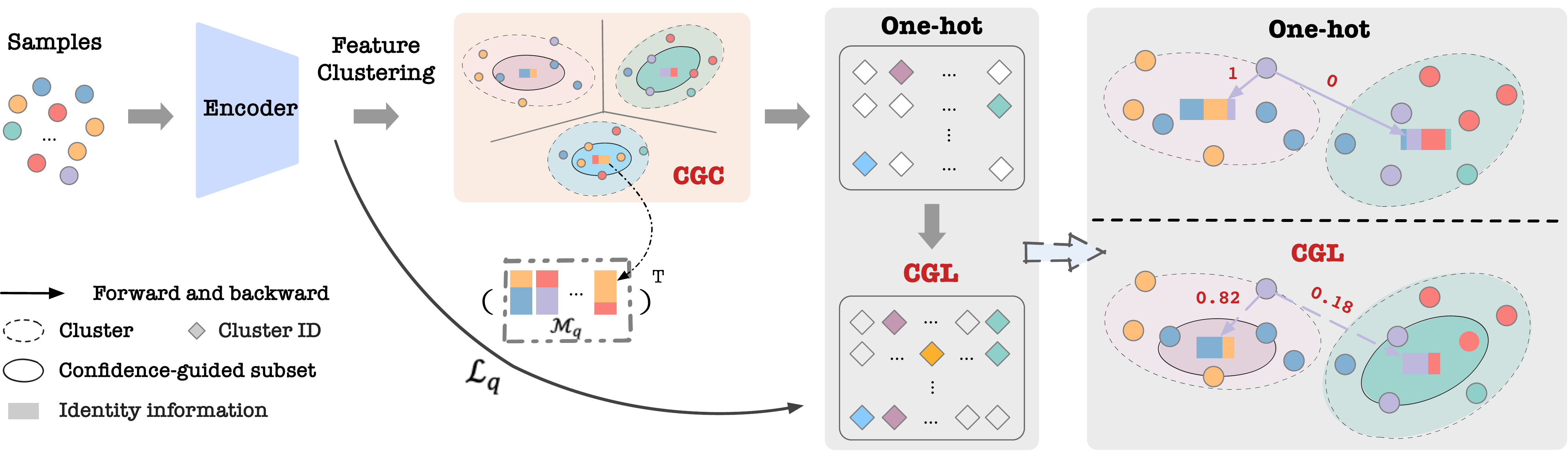}
\caption{{\bf Framework of the proposed method}. At the beginning of each epoch, training samples are clustered by DBSCAN~\cite{ester1996density}. Based on the original clustering result, we select a confidence-guided subset to build our \cgc (CGC). During optimization, samples are encouraged to approach not only the assigned centroid but others where their identity information are potentially embedded via our \cgl (CGL).}
\label{fig:framework}
\vspace{-0.5cm}
\end{figure*}

\section{Methodology}
\label{sec:method}
\subsection{Problem Statement}
Let ${\mathcal{T} = \{x_i\}_{i=1}^{N}}$ denote an unlabeled training dataset, where $x_i$ represents $i$-th image and $N$ is the number of images. 
The USL ReID task aims to train a feature extractor $E_\theta$ in the unsupervised manner, where ReID features $\mathcal{F} = \{f_i\}^{N}_{i=1}$ are derived. 
The identity retrieval during inference is based on such ReID features.
The training scheme of clustering-based USL methods~\cite{ge2020self,dai2021cluster,wang2021camera,zhang2022implicit} alternates between two stages:

\noindent\textbf{Stage I: Clustering.} At the beginning of each epoch, training samples are clustered by DBSCAN~\cite{ester1996density}. Cluster IDs $y_i\in\{1,...,C\}$ serve as one-hot pseudo labels for the network optimization. 
Meanwhile, based on clustering results, a cluster-based memory bank $\mathcal{M}=\{m_i\}^C_{i=1}$ is initialized by cluster centroids that are formulated as,
\begin{equation}
m_i = \frac{1}{|\mathcal{C}|} \sum_{f_i\in \mathcal{C}}f_i,
\label{equ:centroid}
\end{equation}
where $f_i$ represents the feature of $i$-th sample in the cluster $\mathcal{C}$, and $|\mathcal{C}|$ denotes the cluster size.

\noindent\textbf{Stage II: Network Training.} With the obtained pseudo labels, the network is then optimized in a ``supervised'' manner with the training objective, \ie ClusterNCE~\cite{dai2021cluster}, which is formulated as,
\begin{equation}
\centering
\begin{aligned}
\mathcal{L}=-\mathrm{log}\frac{\mathrm{exp}(\Phi (f\cdot m_{+})/\tau )}{\sum_{j=1}^{C}\mathrm{exp}(\Phi (f\cdot m_j)/\tau )},
\end{aligned}
\label{equ:clusternce}
\end{equation} 
where $m_{+}$ refers to the centroid of the cluster that $f$ belongs to, $m_{j}$ represents $j$-th centroid in the memory bank, $\Phi (u\cdot v)$ represents the cosine similarity between vector $u$ and vector $v$, and $\tau$ is the temperature parameter. The memory bank is updated in a momentum manner~\cite{dai2021cluster} as,
\begin{equation}
m_i \leftarrow \mu \cdot m_i + (1 - \mu)\cdot f,
\label{equ:update}
\end{equation}
where $\mu$ is the updating factor and $f$ refers to the feature of instance belonging to $i$-th cluster in the current mini-batch. 

In this paper, we follow the framework of iterative clustering and network training. However, our method, as illustrated in Fig.~\ref{fig:framework}, differs from previous works mainly in two aspects: 1) cluster centroids.
Instead of using all samples to calculate the centroids, we adopt \cgc(CGC) to provide reliable cluster-wise prototypes for feature learning (Sec.~\ref{sec:cgc}), 
and 2) pseudo labels. Apart from the assigned centroid, our \cgl(CGL) encourages instances to approach other centroids where their identity information are potentially embedded (Sec.~\ref{sec:cgl}).

\subsection{Silhouette Score}\label{sec:ss}
To describe the sample-wise clustering confidence, \ie how well a sample fits its cluster, we employ a metric named silhouette score~\cite{rousseeuw1987silhouettes}. The score simultaneously considers two key factors of clustering, \ie tightness and separation.

Formally, for $i$-th data point in cluster $\mathcal{C}_I$, its average distance to other data points within the cluster can be calculated as,
\begin{equation}
    a_i={\frac {1}{|\mathcal{C}_{I}|}}\sum_{i,j\in \mathcal{C}_{I},i\neq j}d(i,j),
\end{equation}
where $d(i,j)$ refers to the distance between $i$-th and $j$-th data points and $|\mathcal{C}_I|$ represents the cluster size.
Similarly, the distance between $i$-th data point and samples belonging to its nearest neighboring cluster $\mathcal{C}_J$ can be denoted as,
\begin{equation}
    b_i=\min_{J\neq I} {\frac {1}{|\mathcal{C}_{J}|}}\sum_{j\in \mathcal{C}_{J}}d(i,j).
\end{equation}
Given the intra-class distance $a_i$ and the minimal inter-class distance $b_i$, the silhouette score $s_i$ is formulated as,
\begin{equation}
    s_i = \frac{b_i - a_i}{max(a_i, b_i)}.
\label{equ:si}
\end{equation}
The silhouette score ranges from $[-1, 1]$. Note that, the score of clusters consisting of a single data point is 0. If an instance has a higher silhouette score, it has a smaller intra-class distance and a large inter-class distance. In other words, it is clustered at a higher confidence~\cite{rousseeuw1987silhouettes}.

\subsection{Confidence-guided Centroids}\label{sec:cgc}
Based on the observation that images with lower silhouette scores (confidence) are generally containing high uncertainty regarding person identity, previous all-sample based cluster centroids are undoubtedly unwise. To remedy the problem, we build \cgc (CGC) with high-confidence images only.

Specifically, the confidence-guided centroid of $i$-th cluster $m_i$ can be formulated as,
\begin{equation}
m_i = \frac{1}{|\mathcal{C}_q|} \sum_{f_i\in \mathcal{C}_q}f_i, \quad \mathcal{C}_q = \{f_i \in \mathcal{C}  | s_i> \delta \}, 
\label{equ:cgc}
\end{equation}
where a confidence-guided subset $\mathcal{C}_q$ is selected from the original cluster $\mathcal{C}$ by a silhouette score threshold $\delta$. 
All confidence-guided centroids are then stored in a confidence-guided memory bank $\mathcal{M}_q = \{m_i\}_{i=1}^C$ for network optimization.

According to Fig.~\ref{fig:sspic}, our \cgc can filter out images that are poor in quality or with cluttered backgrounds at early stages. While at later stages, such centroids effectively exclude some low-confidence samples that possibly belong to other identities.
In summary, the proposed \cgc can provide more reliable cluster-wise prototypes for feature learning.

\subsection{Confidence-guided Pseudo Labels}
\label{sec:cgl}
Another problem of the clustering-based USL methods is that, samples, especially low-confidence ones, very likely carry different identity information with their assigned centroids. Our \cgc also confronts with the problem since only high-confidence samples are included in the formation of centroids, as illustrated in Fig.~\ref{fig:framework}. 
Given the situation, the previous learning scheme, which enforces samples to approach their assigned centroids solely regardless of the identity consistency in-between, is unwise. 
To alleviate the problem, we propose to use \cgl (CGL). 
Such labeling encourages samples to approach not only the assigned centroid but other centroids where their identity information are potentially embedded.

Specifically, we build a distance matrix $\mathcal{D}\in \mathbf{R}^{N\times C}$, where $N$ and $C$ denote the number of samples and clusters at the current epoch, respectively. 
In the paper, clusters consisting of one sample are ignored~\cite{dai2021cluster}. 
As normalized identity features and centroids are adopted, $\mathcal{D}(i,j)$ represents the cosine distance between $i$-th sample and $j$-th confidence-guided centroid. 
Since similar samples are more likely to be scattered in neighboring clusters~\cite{zhang2022implicit}, the identity information of boundary samples is probably embedded in neighboring centroids. 
Therefore, when setting the learning target for samples, neighboring centroids should be assigned with higher confidence while distanced ones should be given lower confidence. 
To this end, a confidence matrix $\mathcal{P}\in \mathbf{R}^{N\times C}$ is obtained by,
\begin{equation}
    \mathcal{P}(i,j) = \frac{p_{i,j}}{\sum_{j=1}^{C} p_{i,j}}, \quad p_{i,j} = \sigma(-\mathcal{D}(i,j)),
\end{equation} 
where $\mathcal{P}(i,j)$ represents the confidence of $j$-th centroid given by $i$-th sample, $\sum_{j=1}^{C}\mathcal{P}(i,j) = 1$, and $\sigma(\cdot)$ is the Sigmoid function. 
By integrating the confidence matrix with the originally assigned one-hot pseudo label $y_i$, the confidence-guided pseudo label of $i$-th sample $\tilde{y}_i$ can be formulated as,
\begin{equation}
    \tilde{y}_i = \beta \cdot {y}_i + (1-\beta)\cdot \mathcal{P}(i),
    \label{equ:cgl}
\end{equation}
where $\beta \in [0,1]$ is the coefficient for the pseudo label refinement.

According to a previous work~\cite{wu2018unsupervised}, the training objective, \ie ClusterNCE, can be considered as a non-parametric classifier, where centroids stored in the memory bank serve as the weight matrix of the classification layer. Therefore, the training objective of our method can be rewritten as, 
\begin{equation}
    \mathcal{L}_q = \frac{1}{N}\sum_{i=1}^N \Big[\ell_{ce} \big(\mathcal{M}_q^T f_i, \tilde{y}_i\big)\Big],
\label{equ:ourloss}
\end{equation}
where $\ell_{ce}$ refers to the cross-entropy loss. 
Compared to Eq.~(\ref{equ:clusternce}), the training objective of our method can be obtained by simply applying two modifications: 
1) replacing the original $\mathcal{M}$ with our confidence-guided memory bank $\mathcal{M}_q$, and 2) replacing the one-hot pseudo label $y_i$ with our confidence-guided one $\tilde{y}_i$.
The training details are presented in Algorithm~\ref{algo}.
\begin{algorithm}[t]
    \textbf{Require:} Unlabeled data with pseudo labels $\mathcal{T}=\{ (x_i, y_i)\}_{i=1}^N$, where $y_i\in\{1, \dots, C\}$\\
    \textbf{Require:} Initialize the backbone encoder $E_\theta$  \\
    \textbf{Require:} Threshold $\delta$ for Eq.~(\ref{equ:cgc}) \\
    \textbf{Require:} Coefficient $\beta$ for Eq.~(\ref{equ:cgl}) \\
    \For{$n$ in $[1, epoch\_num]$}{
        Extracting features $\mathcal{F}$ by $E_\theta$ \\
        Clustering $\mathcal{F}$ into $C$ clusters with DBSCAN \\
        Building CGC dictionary $\mathcal{M}_q$  by  Eq.~(\ref{equ:cgc}) \\
        \For{$m$ in $[1, iteration\_num]$}{
            Sampling a mini-batch from $\mathcal{T}$ \\
            Computing CGL with Eq.~(\ref{equ:cgl}) \\
            Computing loss with Eq.~(\ref{equ:ourloss}) \\
            Updating encoder $E_\theta$ \\ Updating centroids with Eq.~(\ref{equ:update})
        }
    }
    \caption{Pipeline of our method}
    \label{algo}
    \vspace{-0.1cm}
\end{algorithm}

\section{Experiment}
\subsection{Datasets and Evaluation Protocol}
\para{Datasets.} We evaluate our proposed method on \market~\cite{zheng2015scalable}
and \msmt~\cite{wei2018person}. \market includes 32,668 images of 1,501 identities captured by 6 cameras. Among them, 12,936 images of 751 identities are used for training while the resting 19,732 images of 750 identities form the test set. \msmt contains 126,441 images from 4,101 identities captured by 15 cameras. The training set is composed of 32,621 images of 1,041 identities and the test set consists of 93,820 images of 3,060 identities. \msmt is more challenging due to the diversity in backgrounds, illuminations, poses, and occlusions.

\para{Evaluation Protocol.} Following previous methods~\cite{ge2020self,chen2021ice,dai2021cluster,zhang2022implicit}, the mean average precision (mAP)~\cite{bai2017scalable} and the cumulative matching characteristic (CMC)~\cite{zheng2015scalable} top-1, top-5, top-10 accuracies are adopted as evaluation metrics. Note that, there are no post-processing operations, such as reranking~\cite{zhong2017re}, during inference.

\subsection{Implementation Details}
Following previous works~\cite{ge2020self,dai2021cluster,zhang2022implicit}, we adopt ResNet-50~\cite{he2016deep} pre-trained on ImageNet~\cite{deng2009imagenet} as our backbone feature encoder~\cite{dai2021cluster}. All layers after layer-4 are replaced by a generalized mean pooling (GeM)~\cite{radenovic2018fine} layer followed by the batch normalization layer~\cite{ioffe2015batch}. The output 2048-dimensional ReID features are firstly normalized and then used for identity retrieval during inference. Our framework is built upon a state-of-the-art USL method~\cite{dai2021cluster}. For a fair comparison, we follow all experimental settings except for the formation of cluster centroids and the training objectives, as described in Sec.~\ref{sec:method}. The coefficient $\beta$ in Eq. (\ref{equ:cgl}) is empirically set as 0.8 to achieve optimal performances. 

During training, input images are resized to 256$\times$128. We adopt random flipping, cropping, and erasing~\cite{zhong2020random} as data augmentation. Each mini-batch is formed by 16 identities, each with 16 images. Both identity and images are randomly selected from the training set. 
For the optimization, we adopt Adam~\cite{kingma2014adam} optimizer with a weight decay of 0.0005. The learning rate is set to $3.5\times10^{-4}$ initially, and is divided by 10 every 30 epochs. 
We train for a total of 70 epochs on \market~\cite{zheng2015scalable}, and 50 on \msmt~\cite{wei2018person}.

\begin{table*}[]
\small
\scalebox{0.88}{
\begin{tabular}{l|c|cccc|cccc}
\hline
\multicolumn{1}{l|}{\multirow{2}{*}{Method}} & \multicolumn{1}{c|}{\multirow{2}{*}{Reference}} & \multicolumn{4}{c}{\market} & \multicolumn{4}{|c}{\msmt} \\ 
\cline{3-10}
\multicolumn{1}{c|}{} &\multicolumn{1}{c|}{}& \multicolumn{1}{c}{mAP} & \multicolumn{1}{c}{top-1} & \multicolumn{1}{c}{top-5} & \multicolumn{1}{c|}{top-10} & \multicolumn{1}{c}{mAP} & \multicolumn{1}{c}{top-1}  & \multicolumn{1}{c}{top-5} & \multicolumn{1}{c}{top-10}\\ 
\hline
\hline
\multicolumn{2}{l}{\textit{Purely Unsupervised}}\\
\hline
SSL \cite{lin2020unsupervised}&CVPR'20&37.8&71.7&83.8&87.4&-&-&-&-\\
MMCL \cite{wang2020unsupervised} &CVPR'20 &45.5&80.3&89.4&92.3&11.2&35.4&44.8&49.8\\
HCT \cite{zeng2020hierarchical}&CVPR'20 &56.4&80.0&91.6&95.2&-&-&-&-\\
SpCL \cite{ge2020self}&NeurIPS'20 &73.1&88.1&95.1&97.0&19.1&42.3&55.6&61.2\\
JNTL-MCSA~\cite{yang2021joint} & CVPR'21 & 61.7 & 83.9 & 92.3 & - & 15.5 & 35.2 & 48.3 & - \\
GCL \cite{chen2021joint}&CVPR'21&66.8&87.3&93.5&95.5&21.3&45.7&58.6&64.5\\
IICS~\cite{xuan2021intra} & CVPR'21 & 72.9 & 89.5 & 95.2 & 97.0 & 26.9 & 56.4 & 68.8 & 73.4 \\
JVTC+*~\cite{chen2021joint} & CVPR'21 & 75.4 & 90.5 & 96.2 & 97.1 & 29.7 & 54.4 & 68.2 & 74.2 \\
OPLG-HCD~\cite{zheng2021online} & ICCV'21 & 78.1 & 91.1 & 96.4 & 97.7 &26.9 & 53.7 & 65.3 & 70.2 \\
CAP$^{\dagger}$~\cite{wang2021camera}&AAAI'21&79.2&91.4&96.3&97.7&36.9&67.4&78.0&81.4\\
ICE\cite{chen2021ice}&ICCV'21&79.5&92.0&97.0&98.1&29.8&59.0&71.7&77.0\\
ICE$^{\dagger}$~\cite{chen2021ice} &ICCV'21&\textcolor{black}{82.3}&\textcolor{black}{93.8}&\textcolor{black}{97.6}&\textcolor{black}{98.4}&{38.9}&{70.2}&{80.5}&{84.4}\\
Cluster-Contrast~\cite{dai2021cluster}&Arxiv'21& 82.1&92.3&96.7&97.9&27.6&56.0&66.8&71.5\\
PPLR~\cite{cho2022part} & CVPR'22 & 81.5 & 92.8 & 97.1& 98.1 & 31.4 & 61.1 & 73.4& 77.8 \\
PPLR$^{\dagger}$~\cite{cho2022part} & CVPR'22 & 84.4 & 94.3 & 97.8 & 98.6 & 42.2 & 73.3 & 83.5 & 86.5\\
ISE~\cite{zhang2022implicit} & CVPR'22 & 84.7 & 94.0 & 97.8 & 98.8 & 35.0 & 64.7 & 75.5 & 79.4\\
\hline
Cluster-Contrast (*Baseline) & Arxiv'21 & 82.4 & 92.5 & 96.9 & 98.0 & 31.4 & 61.2 & 72.5 & 76.9 \\
\textbf{Baseline+CGC} &-& 84.1 & 93.1 & 97.2& 98.2& 34.1 & 63.1 & 75.0 & 79.0 \\
\textbf{Baseline+CGL} &-& 83.4 & 93.2 & 97.1 & 98.2 & 33.7 & 62.5& 73.9& 78.4 \\
\textbf{Ours}&-&\textbf{85.3}&\textbf{94.2}&\textbf{97.6}&\textbf{98.5}&\textbf{34.6}&\textbf{63.4}&\textbf{74.6}&\textbf{79.3}\\
\hline
\multicolumn{2}{l}{\textit{Supervised}}\\
\hline
PCB \cite{sun2018beyond}&ECCV'18&81.6&93.8&97.5&98.5&40.4&68.2&-&-\\
DG-Net \cite{zheng2019joint}&CVPR'19&86.0&94.8&-&-&52.3&77.2&-&-\\
ICE (w/ ground-truth) \cite{chen2021ice}&ICCV'21&86.6&95.1&98.3&98.9&50.4&76.4&86.6&90.0\\
Our (w/ ground-truth) & - &87.4 & 95.3 & 98.5 & 99.0 & 51.0 & 76.6 & 87.1 & 90.1\\
\hline
\end{tabular}}
\centering
\vspace{-0.cm}
\caption{Comparison of ReID methods on \market and \msmt datasets.
The best USL results without camera information are marked with \textbf{bold}.
$\dagger$ indicates using the additional camera knowledge.
}
\vspace{-0.3cm}
\label{tab:sota}
\end{table*}
\subsection{Comparison with State-of-the-art Methods}
We compare our method with state-of-the-art (SOTA) unsupervised person ReID methods in Table~\ref{tab:sota}. 
Compared with SOTA USL methods, our method outperforms previous ones, except ISE~\cite{zhang2022implicit}, on both benchmarks. Specifically, our method achieves 85.3$\%$ mAP and 94.2$\%$ top-1 accuracy on \market and 34.6$\%$ mAP and 63.4$\%$ top-1 accuracy on \msmt. 
As stated in Sec.~\ref{sec:rw}, existing SOTA methods generally leverage auxiliary information to refine pseudo labels. For example, CAP~\cite{wang2021camera} and ICE~\cite{chen2021ice} leverage the camera information, PPLR~\cite{cho2022part} employs body part predictions, and ISE~\cite{zhang2022implicit} generates extra support samples in the latent space. 
As a departure from the above methods, our method yields SOTA performances by involving internal characteristics, \ie the sample-wise clustering confidence, only.

Additionally, we report the performance of some well-known supervised person ReID methods~\cite{sun2018beyond,zheng2019joint} and unsupervised one~\cite{chen2021ice} under the supervised setting in Table~\ref{tab:sota}. 
Despite the absence of identity labels, our method even outperforms some supervised person ReID methods. Additionally, by replacing the pseudo labels with the ground-truth identity labels provided by datasets, our method outperforms an USL method (ICE~\cite{chen2021ice}), which proves the potential of our framework.

\subsection{Ablation Study}
In this section, we thoroughly analyze the effectiveness of the proposed strategies, \ie \cgc (CGC) and \cgl (CGL). 

\para{Effectiveness of CGC.} We compare models trained with the vanilla all-sample based cluster centroids (``Baseline'') and with the proposed confidence-guided ones (``Baseline + CGC''). The performances are reported in Table~\ref{tab:sota}.
As can be seen, \cgc boost the ReID performance by +1.7$\%$ / +0.6$\%$ on mAP / top-1 accuracy on \market, and +2.7$\%$ / +1.9$\%$ on \msmt. Such improvements reveal the potential of the clustering confidence in the pseudo label refinement.

\begin{figure}
\centering
  \begin{subfigure}{0.9\linewidth}
    \includegraphics[width=0.98\linewidth]{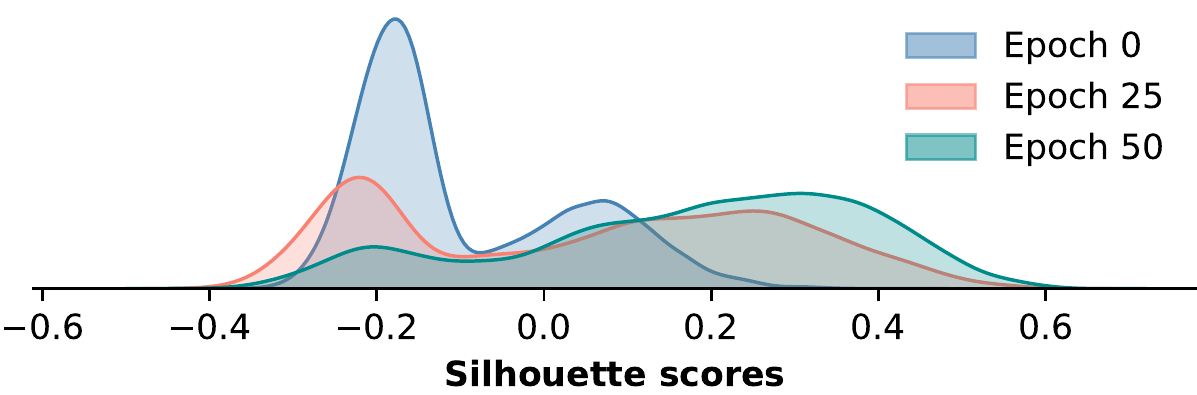}
    \caption{Baseline}
  \end{subfigure}
  \quad
  \begin{subfigure}{0.9\linewidth}
    \includegraphics[width=0.98\linewidth]{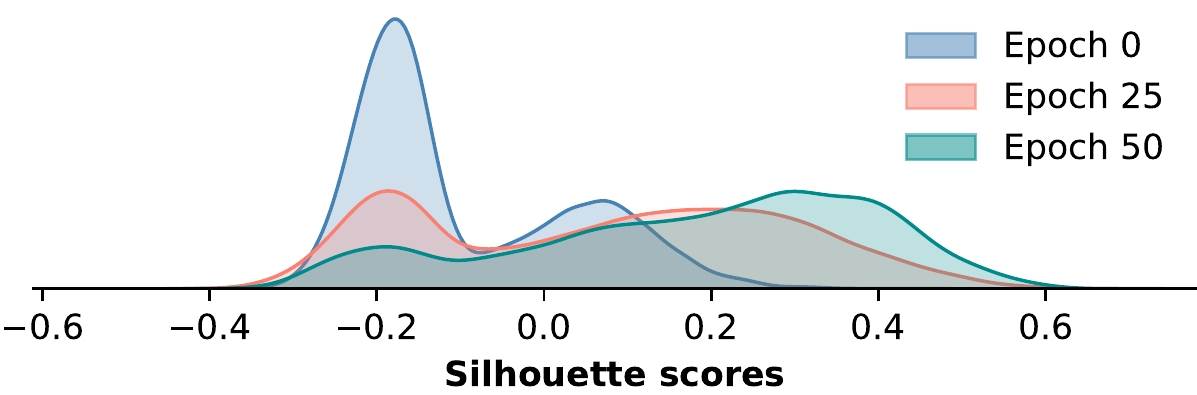}
    \caption{Baseline+CGC}
  \end{subfigure}
  \quad
  \begin{subfigure}{0.9\linewidth}
    \includegraphics[width=0.98\linewidth]{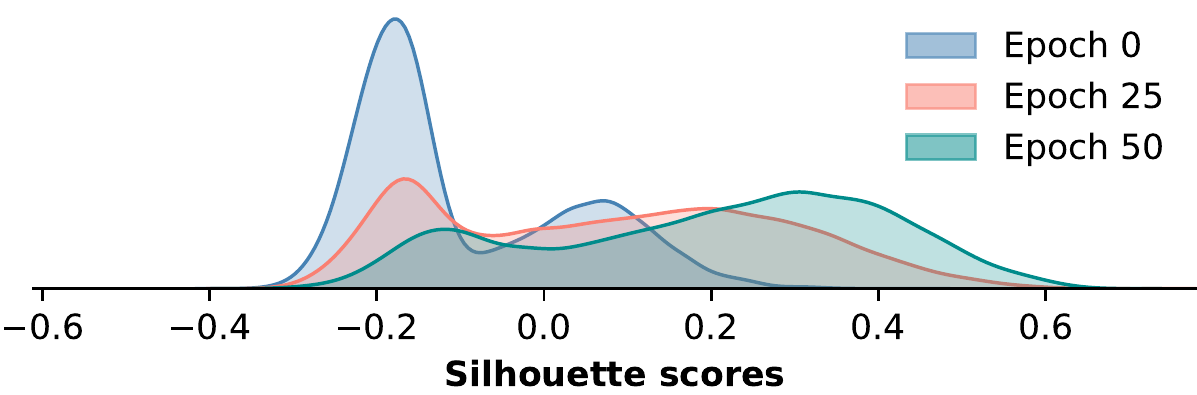}
    \caption{Baseline+CGC+CGL}
  \end{subfigure}
  \caption{Silhouette scores of valid samples (\msmt~\cite{wei2018person}) at different epochs. Comparisons are conducted between (a) baseline model, (b) baseline model with \cgc (CGC), and (c) baseline model with CGC and \cgl (CGL). \textbf{Best viewed in color}.}
\label{fig:slid}
\vspace{-0.3cm}
\end{figure}

To better understand how our \cgc benefit feature learning, we analyze how the sample-wise confidence varies throughout the training process on \msmt. 
Specifically, we visualize the distribution of silhouette scores at different epochs in Fig.~\ref{fig:slid}. 
Note that scores of outliers are excluded.
Several conclusions can be drawn from the comparison between Fig.~\ref{fig:slid}(a) and Fig.~\ref{fig:slid}(b). 
1) As training goes on, the number of valid samples gradually increases, representing as larger areas under the curve. 
2) Starting from the same point (epoch 0), with our \cgc, a noticeable shift towards higher scores can be found at epoch 25. The shift implies CGC can effectively reduce the overall number of low-confidence samples while enhancing high-confidence ones.
3) The advantage remains until the end of training. At epoch 50, the number of high-confidence samples increases, representing by a higher peak closer to 0.4.

\para{Effectiveness of CGL.}
We also compare the baseline model (``Baseline'') and the model trained with \cgl (``Baseline + CGL''). The performances are shown in Table~\ref{tab:sota}. As can be seen, CGL improves mAP and top-1 accuracy by 1.0$\%$ and 0.7$\%$ on \market, by 2.3$\%$ and 1.3$\%$ on \msmt. When both CGC and CGL are employed during training, improvements are +2.9$\%$ and +1.7$\%$ on \market, and +3.2$\%$ and +2.2$\%$ on \msmt.

In terms of the sample-wise clustering confidence, we visualize the distribution of silhouette scores in Fig.~\ref{fig:slid}(c), when CGL is applied during training. 
Compared to the model trained without CGL (Fig.~\ref{fig:slid}(b)), CGL further pushes the score towards a higher value at both epoch 25 and epoch 50. 
Less low-confidence samples during training implies our CGL contributes to better clustering.
In summary, the above qualitative and quantitative results prove the proposed scheme can boost performance by enhancing the sample-wise clustering confidence. 

\subsection{Parameter Analysis}
\para{Threshold $\delta$ in CGC.} \label{sec:zeta}
To obtain the optimal threshold $\delta$ in Eq.~(\ref{equ:cgc}) for the proposed \cgc (CGC), three types of threshold selection strategies are explored, \ie linear, dynamic and constant, respectively. 
For the former two strategies, the threshold gradually increases as training goes on. 
The constant strategy employs a fixed threshold throughout the training process. 

Specifically, the linear strategy updates the threshold by $\delta_{t}=\delta_0 * t/T + \epsilon$, where $\delta_0$ limits the range of threshold and $\epsilon$ is the offset. 
In the paper, we set $\delta_0=0.2$ and $\epsilon=-0.1$. $t$ and $T$ denote the current epoch and the overall number of epochs, respectively. 
In terms of the dynamic strategy, the threshold is updated by $\delta = \delta_0 * tanh(0.1*(t-T/2))$. We set $\delta_0 = 0.1$ to achieve $\delta\in [-0.1,0.1]$, which is the same as the linear strategy. 
The range is set empirically in the consideration of the image quality and the distribution of silhouette scores (see Fig.~\ref{fig:slid}). 
Apart from the varying threshold, we conduct the constant strategy by fixing the threshold as $\{-0.1, 0, 0.1\}$ respectively. 
The comparisons between model performances with different strategies are reported in Table~\ref{tab:thre}. 
The best performance is achieved when adopting the linear strategy for \market and applying a fixed threshold $\delta=0$ on \msmt. 
The optimal settings are employed in all experiments.

\begin{table}
\centering
\resizebox{0.45\textwidth}{!}{
    \begin{tabular}{c|c|c|cc|cc}
        \toprule
        \multirow{2}{*}{Method} & \multirow{2}{*}{Strategy} & \multirow{2}{*}{$\delta$}& \multicolumn{2}{c|}{\market} & \multicolumn{2}{c}{\msmt} \\
        \cline{4-7}
         & & & mAP & top-1 & mAP & top-1 \\ 
        \hline
        Baseline & - & - & 82.4 & 92.5 & 31.4 & 61.2 \\
        \hline
        \multirow{5}{*}{Ours} & Linear & - & \textbf{85.3} & \textbf{94.2} & 33.6 & 63.0\\
         & Dynamic & - & 84.9 & 93.9 & 33.0 & 62.8  \\
        \cline{2-7}
         & \multirow{3}{*}{Constant} & -0.1 & 83.5 & 93.4 & 32.7 & 62.8 \\
         &  & 0 & 84.9 & 94.0 & \textbf{34.6} & \textbf{63.4} \\
         &  & 0.1 & 84.0 & 93.3 & 34.0 & 63.2 \\
        \bottomrule
    \end{tabular}
    }
\caption{Comparison of threshold selection strategies of \cgc(CGC) on benchmark datasets.}
\label{tab:thre}
\vspace{-0.3cm}
\end{table}

\begin{figure}
\centering
  \begin{subfigure}{0.48\linewidth}
    \includegraphics[width=0.99\linewidth]{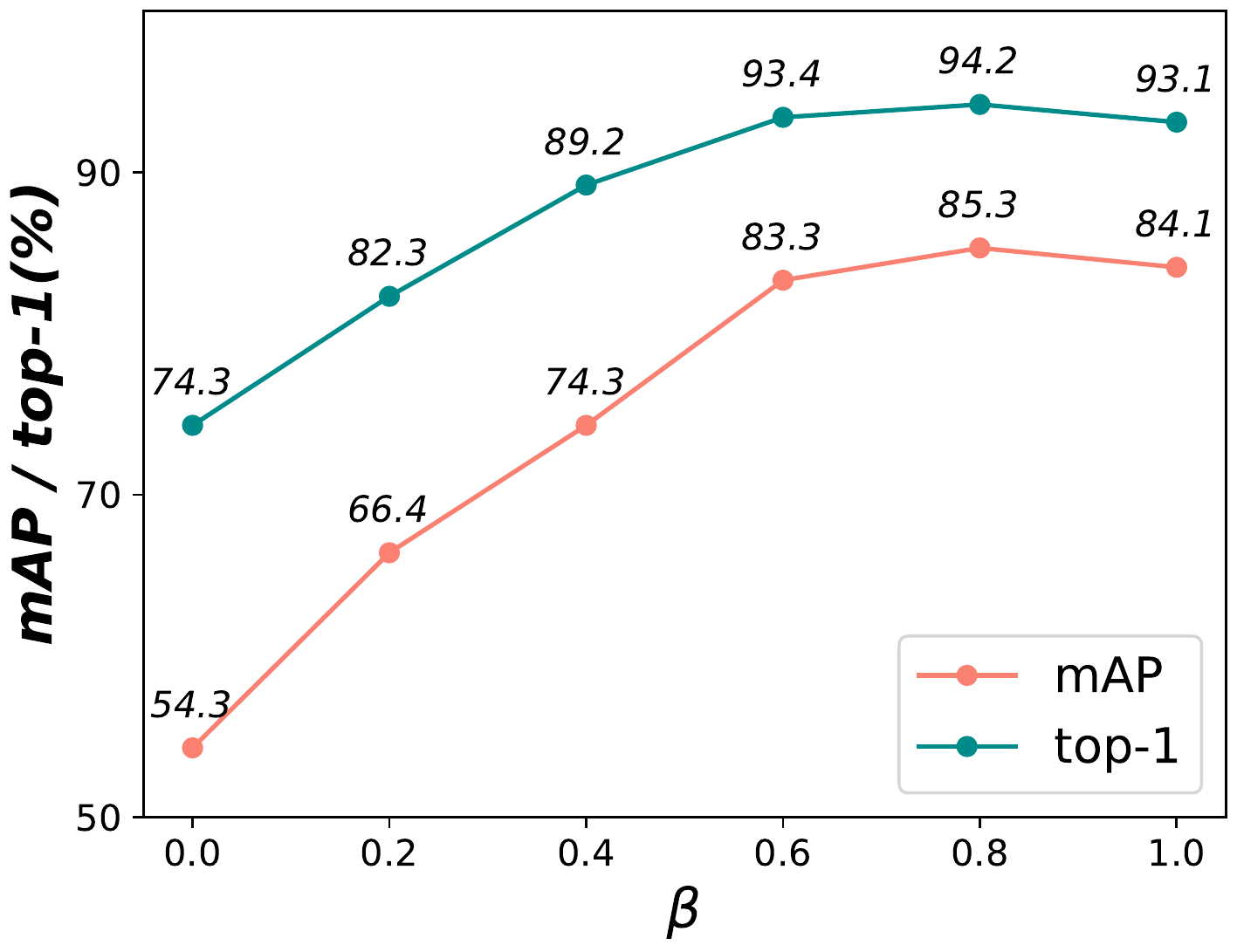}
    \caption{\market}
  \end{subfigure}
  \hspace{.6mm}
  \begin{subfigure}{0.48\linewidth}
    \includegraphics[width=0.99\linewidth]{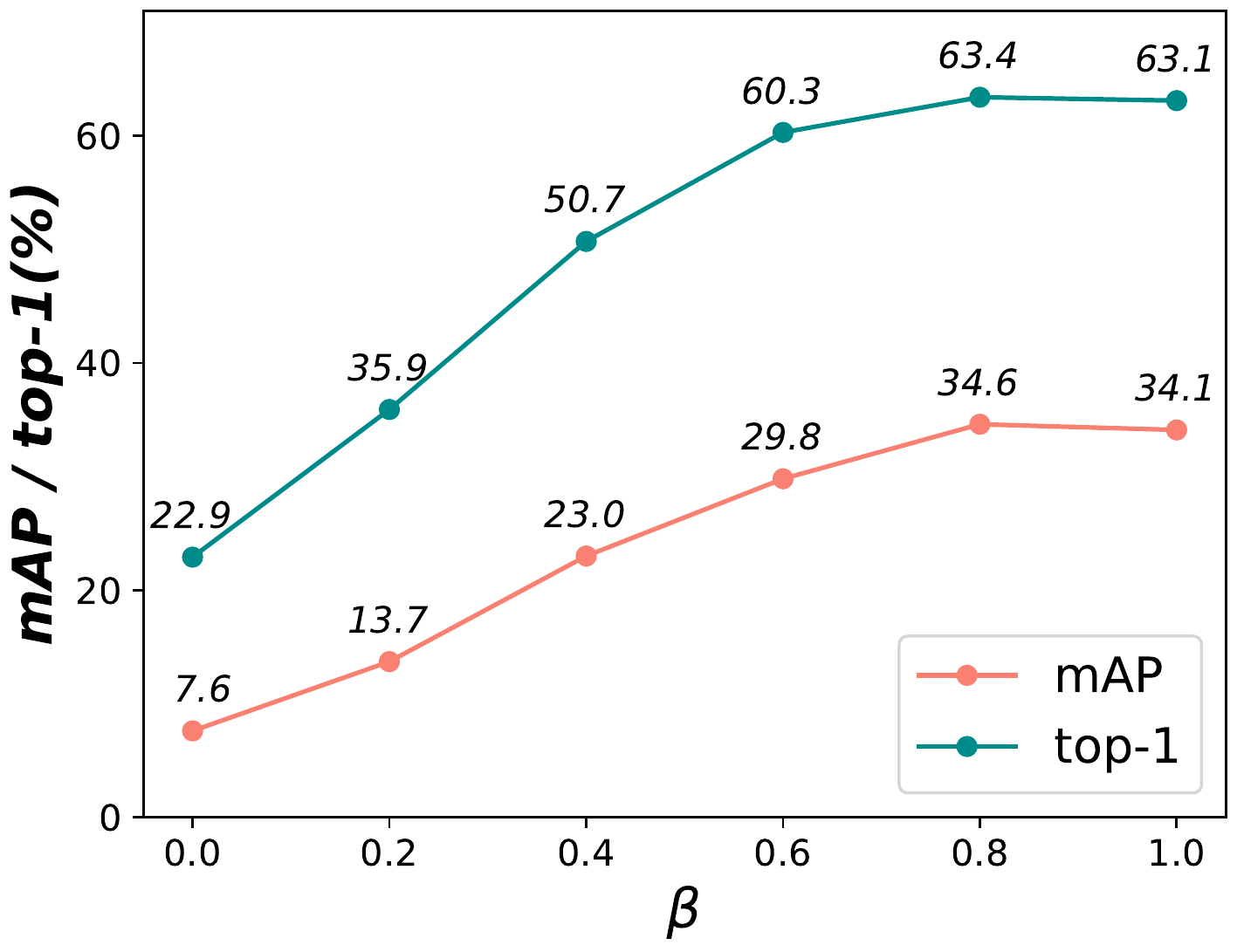}
    \caption{\msmt}
  \end{subfigure}
  \vspace{-0.2cm}
  \caption{Comparison of coefficient $\beta$ in \cgl(CGL) on (a) \market and (b) \msmt.}
  \label{fig:beta}
  \vspace{-0.2cm}
\end{figure}

\para{Coefficient $\beta$ in CGL.}
To analyze the impact of the coefficient $\beta$ in the proposed \cgl (CGL), we tune the value of parameter $\beta$ from 0 to 1 while keeping others fixed.
According to Eq.~(\ref{equ:cgl}), when $\beta$ is set to 0 or 1, our method decomposes down to using the confidence matrix or the one-hot pseudo label exclusively during training. 
The results on two benchmarks are illustrated in Fig.~\ref{fig:beta}. 
As shown, as $\beta$ increases from 0 to 0.8, both mAP and top-1 accuracy increase. A slight performance drop can be found when increasing $\beta$ from 0.8 to 1. 
To achieve the best performance, we set $\beta=0.8$ for all experiments.

\subsection{More Discussions}\label{sec:diss}
\begin{figure}[htbp]
\centering
  \begin{subfigure}{0.94\linewidth}
    \includegraphics[width=0.99\linewidth]{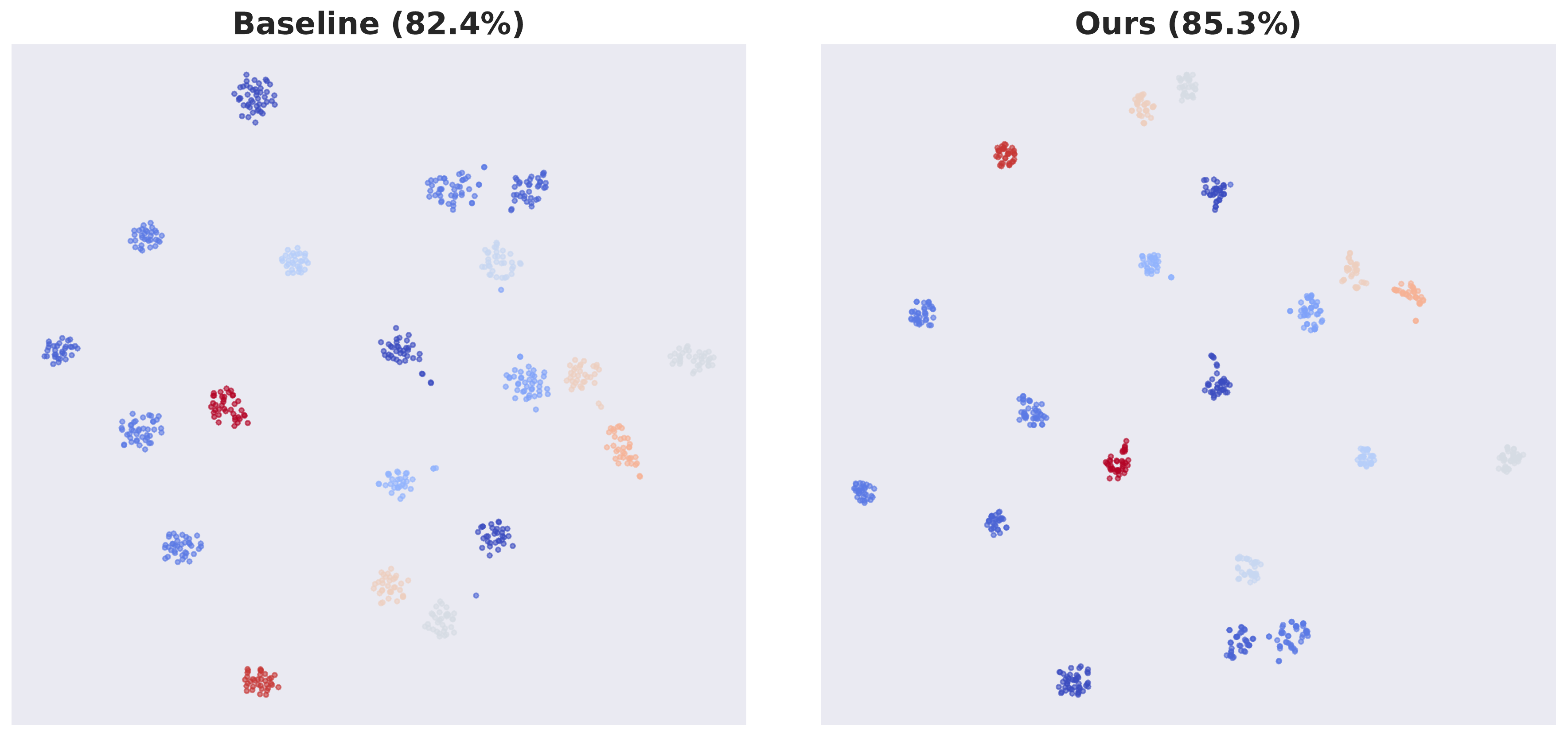}
    \caption{\market}
  \end{subfigure}
  \quad
  \begin{subfigure}{0.94\linewidth}
    \includegraphics[width=0.99\linewidth]{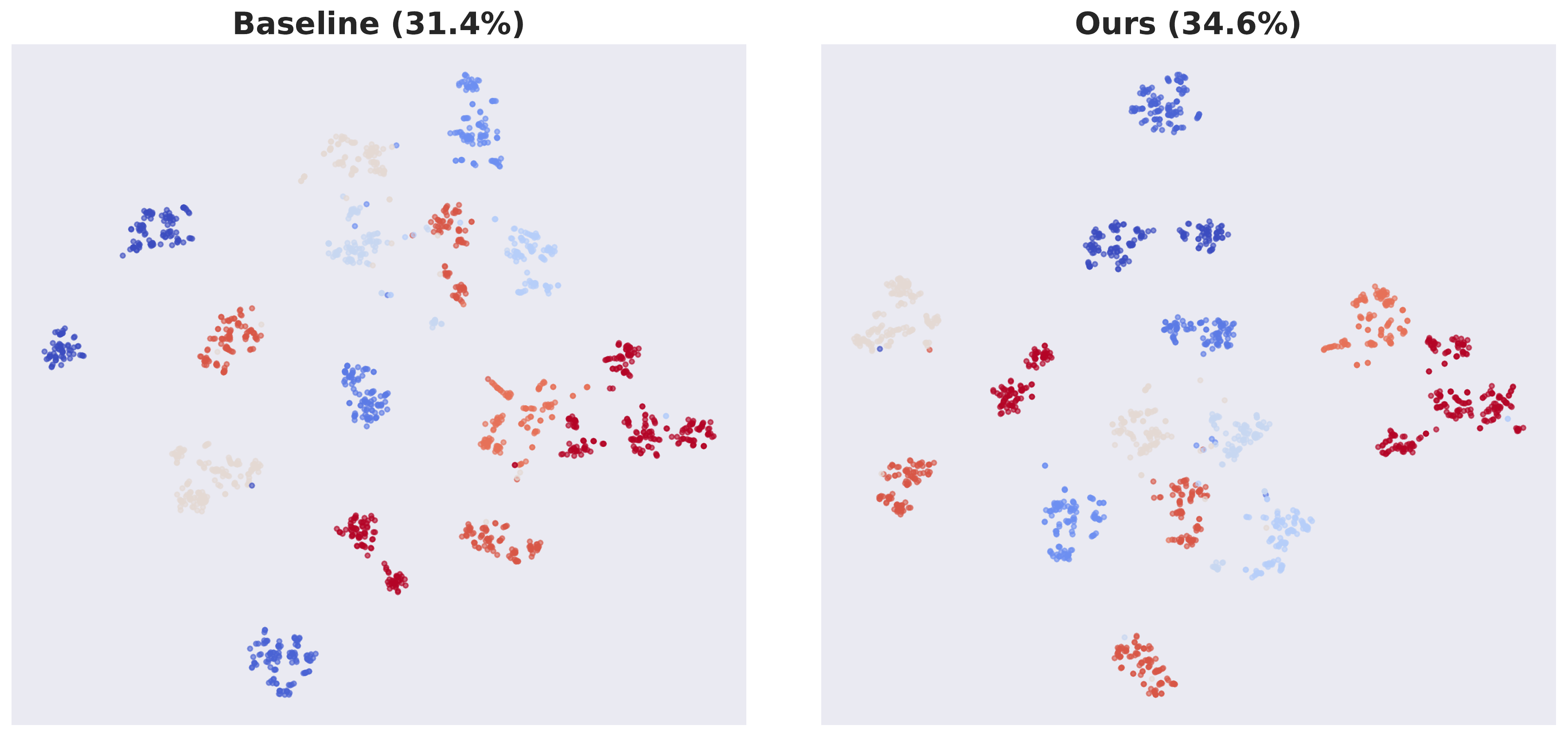}
    \caption{\msmt}
  \end{subfigure}
  \caption{Visualization of the identity feature distribution via t-SNE~\cite{van2008visualizing} on (a) \market and (b) \msmt. For each group, features are derived by the baseline model (left) and the model trained with the proposed confidence-guided centroids (CGC) and pseudo labels (CGL) (right), respectively. Model performances (mAP) are also denoted. Different identities are denoted by different colors. \textbf{Best viewed in color}.}
  \label{fig:pidtsne}
  \vspace{-0.4cm}
\end{figure}

\para{Identity Feature Distribution.} To better understand the advantages of the proposed strategies, we visualize the distribution of identity features via t-SNE~\cite{van2008visualizing}.
Specifically, 20 identities are randomly selected from \market~\cite{zheng2015scalable} and \msmt~\cite{wei2018person}, respectively. Features of selected identities are extracted by the baseline model and our model is trained with \cgc (CGC) and \cgl (CGL). 
The distribution of identity features is illustrated in Fig.~\ref{fig:pidtsne}. 
As can be seen, due to the vast variety in camera views, backgrounds, and poses, the feature distribution of \msmt is more chaotic than that of \market. 
Despite such challenges, with the aid of the proposed strategies, features of the same identity are distributed more compactly while features of different identities are further separated.

\para{Identity Consistency Score.}
The current learning scheme enforces samples to approach their assigned cluster centroids, where their identity information are embedded.
However, the existence of noisy labels will lead samples to ``wrong'' centroids.
It is especially problematic for low-confidence samples, \ie boundary samples, because they can be closer to other centroids than the assigned ones. 

To investigate the problem, we conduct an experiment on \msmt to analyze how much the identity information of boundary samples can be presented in the assigned centroids, \ie the identity consistency in-between.
Specifically, we select clusters whose size is over 100 at each epoch. For each cluster, samples whose silhouette scores rank at the bottom 5$\%$ are empirically marked as boundary samples.
Formally, let $\mathcal{C} = \{(x_i, g_i)\}_{i=1}^{N_c}$ denote a cluster with $N_c$ samples, where $g_i$ refers to the ground-truth identity label provided by the dataset. An identity set $\mathcal{G} = \{g_k\}_{k=1}^M$ is then constructed by overall $M$ identities occurring in the cluster.
Following the formation of vanilla all-sample based cluster centroids (Eq.~(\ref{equ:centroid})), the identity information embedded in the centroid can be obtained by linearly integrating all identities within the cluster via weights $\mathcal{Q} = \{q_k\}_{k=1}^{M}$, where $q_k$ is obtained by $q_k = \frac{1}{| \mathcal{C} |} { \sum_{g_i \in \mathcal{C}} \mathbbm{1} \{g_i = g_k\} }$.
$| \mathcal{C} |$ denotes the cluster size. 
$\mathbbm{1} \{g_i = g_k\}$ equals to 1 when $g_i = g_k$, otherwise 0. 
Then, the identity consistency score (ICS) between boundary samples and the cluster centroid of $\mathcal{C}$ can be calculated as, $ICS = \frac{1}{N_c} \sum_{g_i \in \mathcal{C}} {q_k} \cdot \mathbbm{1} \{g_i = g_k\}.$

\begin{figure}
\centering
\includegraphics[width=.6\linewidth]{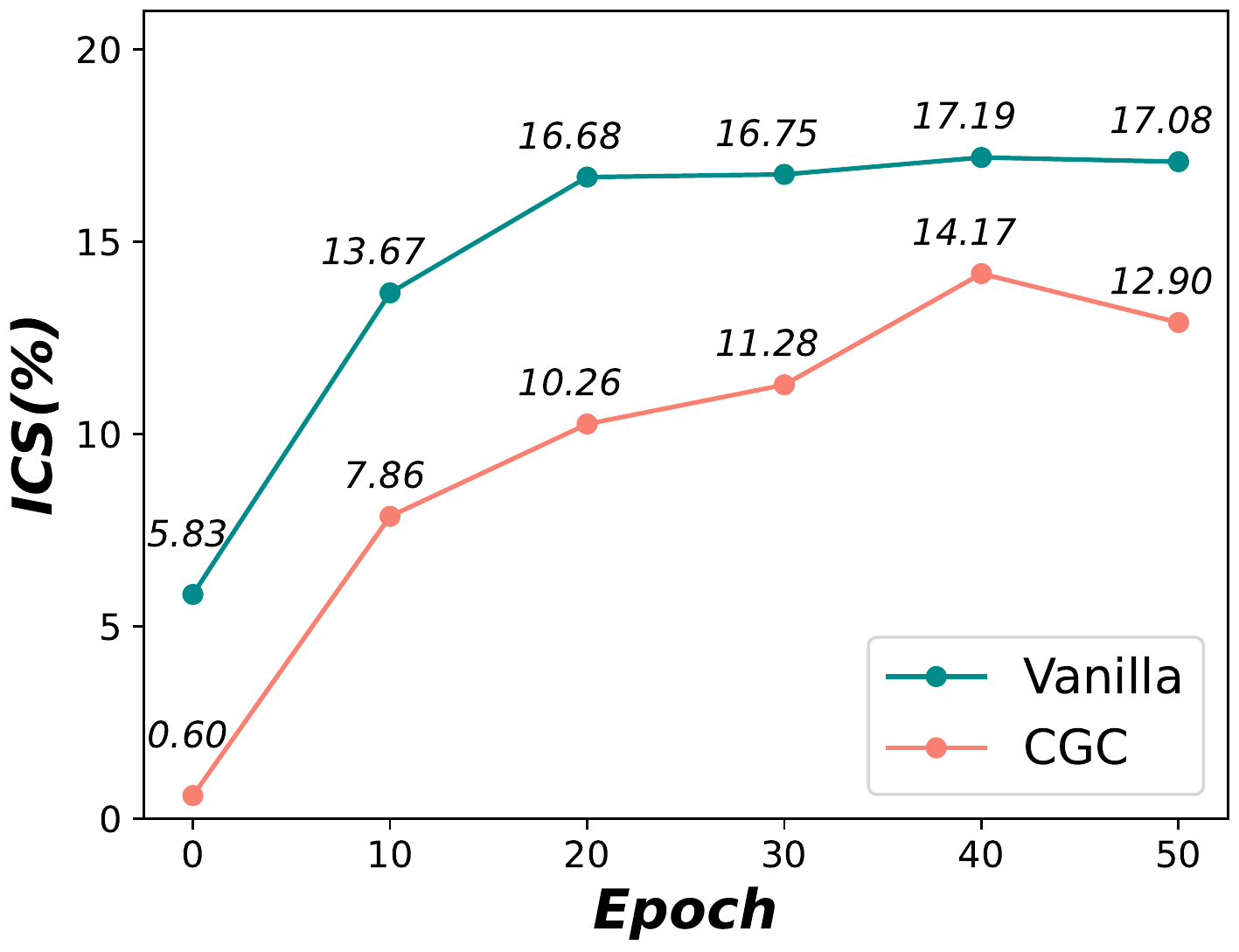}
\vspace{-0.1cm}
\caption{Identity consistent score (ICS) of boundary samples at different epochs. Vanilla and CGC refer to the previous all-sample based cluster centroids and the proposed \cgc, respectively.}
\label{fig:ratio}
\vspace{-0.5cm}
\end{figure}

Similar to the vanilla scheme, ICS of our \cgc (CGC) scheme can be computed by simply replacing $\mathcal{C}$ with the confidence-guided subset $\mathcal{C}_q$ during the computation of the weight $q_k$.
Since low-confidence samples are filtered out in the formation of \cgc, the identity set $\mathcal{G}$ only includes identities of samples with high confidence scores. 
We compare the average ICS throughout the training with vanilla all-sample based cluster centroids and the proposed confidence-guided ones, and obtain the curves in Fig.~\ref{fig:ratio}.

For the vanilla scheme, only 5.83$\%$ boundary samples carry the same identity information with their assigned centroid at the beginning. 
Although the ratio gradually climbs to 17.19$\%$, a large proportion of boundary samples (over 80$\%$) still are pushed to centroids where their identity information are rarely presented. Unfortunately, the problem has been aggravated by \cgc, where the ratio achieves 14.17$\%$ at most.
The low identity consistency scores point out the seriousness of the problem and validates the necessity of our \cgl. 

\section{Conclusion}
This paper focused on the pseudo label refinement for clustering-based unsupervised person ReID, which aims to alleviate the pseudo label noise brought by imperfect clustering results.
Instead of relying on auxiliary information such as camera IDs, body parts, or generated samples, we refined pseudo labels with internal characteristics, \ie the sample-wise clustering confidence.
Specifically, we proposed to use \cgc (CGC) to provide reliable cluster-wise prototypes for feature learning, where low-confidence instances are filtered out during the formation of centroids.
Additionally, targeting at the problem that a large proportion of samples are pushed to ``wrong'' centroids, we propose to use \cgl (CGL). 
Such labeling enables samples to approach not only the assigned centroid but other clusters where their identities are potentially embedded.
With the aid of CGC and CGL, our method yields comparable performances with, or even outperforms, state-of-the-art pseudo label refinement works that largely leverage auxiliary information.

\para{Limitations and Broader Impact.}
Although we conducted multiple threshold strategies in the paper, the range is selected empirically. We are interested in exploring adaptive thresholds in the future.
Despite that our method did NOT leverage either identity labels or auxiliary information, it may still involve a concern for human privacy during the data collection.
Therefore, the legal utilization of person ReID data should be regulated strictly to avoid ethical issues.

{\small
\bibliographystyle{ieee_fullname}
\bibliography{my}
}

\end{document}